\documentclass[journal]{IEEEtran}
\usepackage[T1]{fontenc}% optional T1 font encoding
\usepackage{graphicx}
\usepackage{times}
\usepackage{helvet}
\usepackage{courier}
\usepackage{amsmath}
\usepackage{algorithm}
\usepackage{algorithmic}
\usepackage{csquotes} 
\usepackage{color}
\usepackage{paralist}
\usepackage{amssymb}
\usepackage{indentfirst}
\usepackage{subfigure}
\usepackage{float}
\usepackage{multirow}
\usepackage{cite}
\usepackage{mathrsfs}

\usepackage{mathrsfs} 
\usepackage{amsfonts}
\usepackage{todonotes}
\usepackage{pgfplots} %引用包
\usepackage{epstopdf}
\usepackage{epsfig}
\pgfplotsset{compat=newest}
\usepackage{color}
\definecolor{forestgreen}{RGB}{0,139,69}

\usepackage{xcolor}
\definecolor{citecolor}{HTML}{0071bc}
\usepackage[colorlinks, linkcolor=red,  anchorcolor=blue, citecolor=citecolor]{hyperref} 
\definecolor{SeaGreen4}{RGB}{0,205,102} 
\definecolor{SlateBlue}{RGB}{106,90,205} 
\definecolor{DarkRed}{RGB}{178,34,34} 
	
\usepackage[switch]{lineno}
\usepackage{textcomp,booktabs}
\usepackage{amssymb}% http://ctan.org/pkg/amssymb
\usepackage{pifont}% http://ctan.org/pkg/pifont

\usepackage{makecell}

\usepackage{colortbl}
\definecolor{mygray}{gray}{.9}
\definecolor{mypink}{rgb}{.99,.91,.95}
\definecolor{mycyan}{cmyk}{.3,0,0,0}

\begin{document}

\title{ Revisiting Heat Flux Analysis of Tungsten Monoblock Divertor on EAST using Physics-Informed Neural Network}

\author{Xiao Wang, \emph{Member, IEEE}, Zikang Yan, Hao Si, Zhendong Yang*, Qingquan Yang*, \\ Dengdi Sun, Wanli Lyu, Jin Tang 
\thanks{$\bullet$ Xiao Wang, Zikang Yan, Hao Si, Dengdi Sun, Wanli Lyu, and Jin Tang are with the School of Computer Science and Technology, Anhui University, Hefei 230601, China. (email: xiaowang@ahu.edu.cn)} 
\thanks{$\bullet$ Zhendong Yang is with Tongling University, Tongling 244000, China (email: dongyz@tlu.edu.cn)} 
\thanks{$\bullet$ Qingquan Yang is with the Institute of Plasma Physics, Chinese Academy of Sciences, Hefei, China. (email: yangqq@ipp.ac.cn)} 
\thanks{* Corresponding Author: Zhendong Yang, Qingquan Yang} 
}

\markboth{ IEEE Transactions on ***, 2025 } 
{Shell \MakeLowercase{\textit{et al.}}: Bare Demo of IEEEtran.cls for IEEE Journals}

\maketitle

% As a general rule, do not put math, special symbols or citations in the abstract or keywords.
\begin{abstract}
Estimating heat flux in the nuclear fusion device EAST is a critically important task. Traditional scientific computing methods typically model this process using the Finite Element Method (FEM). However, FEM relies on grid-based sampling for computation, which is computationally inefficient and hard to perform real-time simulations during actual experiments. Inspired by artificial intelligence-powered scientific computing, this paper proposes a novel Physics-Informed Neural Network (PINN) to address this challenge, significantly accelerating the heat conduction estimation process while maintaining high accuracy. Specifically, given inputs of different materials, we first feed spatial coordinates and time stamps into the neural network, and compute boundary loss, initial condition loss, and physical loss based on the heat conduction equation. Additionally, we sample a small number of data points in a data-driven manner to better fit the specific heat conduction scenario, further enhancing the model's predictive capability. We conduct experiments under both uniform and non-uniform heating conditions on the top surface. Experimental results show that the proposed thermal conduction physics-informed neural network achieves accuracy comparable to the finite element method, while achieving $\times$40 times acceleration in computational efficiency. 
% The dataset and source code will be made publicly available upon paper acceptance.
The dataset and source code will be released on \url{https://github.com/Event-AHU/OpenFusion}
\end{abstract}


\IEEEpeerreviewmaketitle

\section{Introduction}

\begin{figure*}[t] % 尝试将图片置于页面顶部（跨双栏）
\centering % 图片整体居中
\includegraphics[width=1\textwidth]{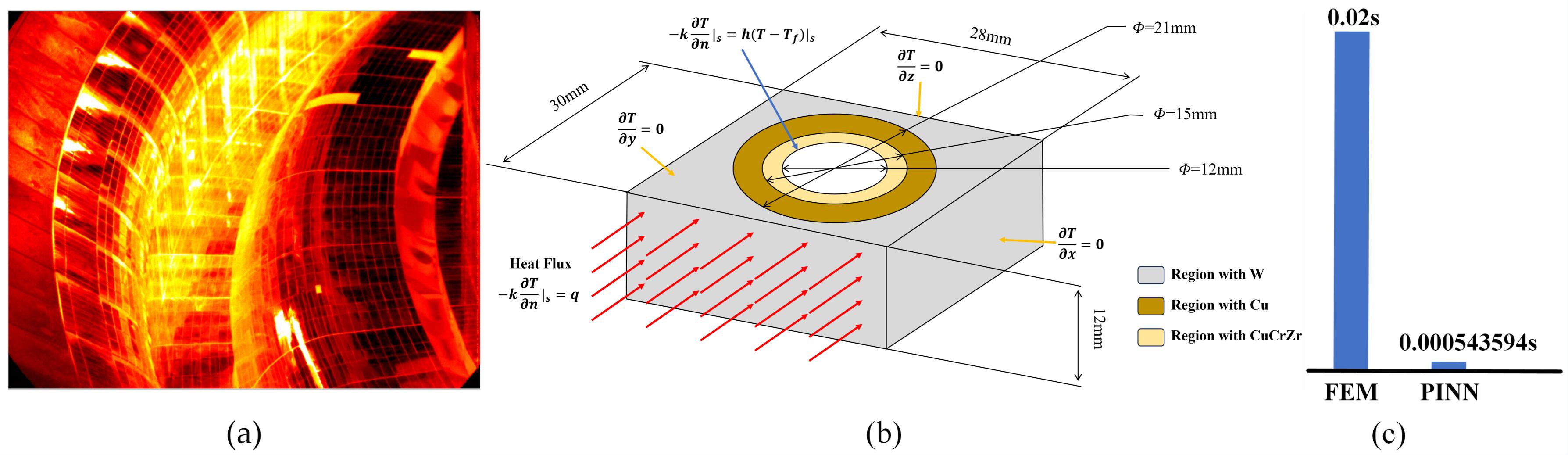} % 缩放图片至原宽度的 80%
\centering
\caption{(a) Baking image of a flat plate divertor. (b) Schematic diagram of a single water cooling device, with materials layered from the outermost to the innermost as W, OFHC-Cu, CuCrZr, and condensate water at the innermost layer. (c) Comparison diagram of time required for prediction at 5773 points by the Finite Element Method (FEM) and Physics-Informed Neural Network (PINN) method.} 
\label{fig::firstIMG}
\end{figure*}

\IEEEPARstart{N}{uclear} Fusion (NF) is a process in which two light atomic nuclei combine to form a heavier nucleus, releasing a tremendous amount of energy. It is the same reaction that powers the Sun and other stars. As a clean energy source, nuclear fusion offers significant advantages, including abundant fuel supply, extremely high energy density, environmental friendliness with no greenhouse gas emissions, and inherent safety. Therefore, nuclear fusion is considered a crucial pathway toward achieving sustainable clean energy for humanity and has become a major scientific research direction, receiving substantial investment and development efforts from countries around the world. Among the various approaches, the tokamak-based technology represents a leading research focus. Notable devices include China's EAST (Experimental Advanced Superconducting Tokamak)~\footnote{\url{http://east.ipp.ac.cn/}} and HL-3~\footnote{\url{https://www.swip.ac.cn/swip/english66/tokamaks/1366463/index.html}}, as well as the international fusion project ITER~\footnote{\url{https://www.iter.org/}}.

Due to the complexity of the device and its experimental tasks, numerous diagnostic systems have been developed around nuclear fusion to assist researchers in understanding and analyzing experimental parameter settings and related results~\cite{wang2024multi, ma2024exploiting}.  
Among these, heat flux estimation~\cite{yang2020FEMHEATEAST, shi2017heat, shi2018study} on the divertor is a particularly important task for the study of the characteristics of energy transport in controlled fusion plasma~\cite{chen2023development}, as shown in Fig.~\ref{fig::firstIMG} (a). Traditional methods to handle this problem are widely used Finite Element Method (FEM)~\footnote{\url{https://en.wikipedia.org/wiki/Finite_element_method}}, which is a numerical method extensively applied in engineering and scientific fields to solve complex physical problems. It discretizes continuous physical systems into a finite number of elements and approximates solutions to partial differential equations through these elements. For example, Yang et al.~\cite{yang2020FEMHEATEAST} develop a 3D FEM heat flux calculation method with full consideration of the monoblock geometry of the upper divertor. Despite its effectiveness, the high computational cost of the FEM framework makes it difficult to deploy in devices for real-time and efficient control, as shown in Fig.~\ref{fig::firstIMG} (c). Therefore, developing a highly efficient scheme for estimating heat flux remains an urgent problem to be solved.

On the other hand, the explosive growth of artificial intelligence technology in recent years, particularly breakthroughs in deep neural networks and large foundation models~\cite{wang2023MMPTMSurvey, wang2025xihefusion}, has also inspired a shift in the research paradigm for nuclear fusion science. An increasing number of researchers have begun incorporating deep neural networks into the computation of physical formulas, leading to the development of a series of Physics-Informed Neural Networks (PINNs)~\cite{raissi2019physics}. Specifically, 
Wang et al.~\cite{wang2024physics} developed a Physics-Informed Neural Network based on the dynamic behavior of battery degradation and degradation trends to accurately and stably estimate the State of Health (SOH) of batteries; 
Jarolim et al.~\cite{jarolim2023probing} proposed a coronal magnetic field extrapolation method that integrates observational data and a physically constrained force-free magnetic field model into a neural network, aiming to analyze the solar coronal magnetic field; 
Jang et al.~\cite{jang2024grad} trained neural networks by relying solely on the physical constraints inherent in the equations themselves to address the crucial magnetohydrodynamic (MHD) equilibrium problem in plasma physics. A brief introduction to the working principle of PINNs can be found in Fig.~\ref{fig:2D_heat_conduction}.

Inspired by these works, in this paper, we revisit the heat flux analysis of the tungsten monoblock divertor on EAST using physics-informed neural networks. The key insight of our model lies in the fact that the core of the heat flux estimation problem is modeling heat conduction under different materials, and heat conduction can be explicitly expressed by well-defined physical formulas. By combining deep neural networks with the physical equations of heat conduction to predict heat flux, we can fully leverage the prior knowledge of physics and the powerful fitting capabilities of data-driven neural networks. 
Specifically, as shown in Fig.~\ref{fig::framework}, we first sample the points for boundary conditions, initial conditions, and residuals, from the defined spatio-temporal domain of the physical model, and extend each point into a continuous region to address the inaccurate solution caused by discrete points. Then, the normalization is conducted on these data to eliminate dimensional and range discrepancies. Considering different materials in our case, we design three fully-connected sub-networks to address these different domains. We resort to the continuity conditions to model the correlations between different sub-networks. In addition to the physical-aware loss functions, we also introduce a small amount of supervised data to augment the optimization of our PINNs. Extensive experiments in diverse scenarios fully validated the effectiveness of our proposed strategy for the heat flux estimation.

To sum up, the main contributions of this paper can be summarized as the following three aspects:

1). We propose a novel deep learning paradigm jointly driven by physics constraints and sparse data, offering a new artificial intelligence alternative for heat flux analysis of the tungsten monoblock divertor on EAST. 

2). We propose a new physics-informed neural network to replace the FEM for the heat flux analysis problem, termed HFPINN. It uses three sub-networks for different materials, linked by a consistency loss, and applies regional optimization to avoid poor local solutions in PINNs.  

3). Extensive experiments under both uniform and non-uniform upper surfaces demonstrate that our sparse data-driven PINN model can achieve accuracy close to that of the FEM method, while significantly accelerating the computation process.

\section{Related Works}

In this section, we focus on reviewing the related works on Numerical Method Reconstruction of EAST Divertor, Physics-Informed Neural Network, and Heat Conduction. More related works can be found in the following surveys~\cite{cai2021PINNSurvey, karniadakis2021physics} and paper list~\footnote{\url{https://github.com/Event-AHU/PINN_Paper_List}}.

\subsection{Numerical Method Reconstruction of EAST Divertor}
Partial Differential Equations (PDEs) play a critical role in science and engineering. Many complex physical systems cannot be solved analytically but are instead modeled using PDEs. Computational simulations for PDE systems have long been a core focus of research, with numerical methods such as the Finite Difference Method (FDM)~\cite{causon2010introductory}, the Finite Element Method (FEM)~\cite{reddy1993introduction}, and the Finite Volume Method (FVM)~\cite{eymard2000finite} serving as powerful tools. Power loading on the first wall constitutes a critical challenge in fusion devices (e.g., ITER~\cite{brunner2011comparison}, CFETR~\cite{wan2017overview}). During steady-state operations, a large amount of heat flux concentrates on narrow regions of the divertor target plates. Transient events such as Edge Localized Modes (ELMs) and disruptions cause dramatic spikes in heat flux. This immense localized thermal load severely damages Plasma-Facing Components (PFCs), jeopardizing device safety and stable operation. Consequently, precise calculation of the power flux onto the divertor has become an urgent task to ensure fusion device reliability and enable effective heat flux control. Shi et al.~\cite{shi2017heat} used a two-dimensional finite element analysis code named DFLUX to calculate the heat flux distribution and its peak value on the target plate, and investigated the critical influence of comparing the two hybrid heating methods (LHW+NBI and LHW+ICRH) on the heat load on the divertor target plate during H-mode discharges. Using similar methods, Shi et al.~\cite{shi2018study} also compared and analyzed the impacts of pure LHW heating versus the LHW+NBI hybrid heating method on the heat load of the divertor target plate. Since 2014, EAST has been upgraded to a water-cooled W/Cu monoblock, and therefore, the original two-dimensional(2D) heat calculation code does not apply to the heat flux calculation on the upper divertor due to the asymmetry toroidal geometry. Thus, Yang et al. ~\cite{yang2020FEMHEATEAST} developed a 3D FEM heat flux calculation method with full consideration of the monoblock geometry of the upper divertor. Based on surface temperature data obtained by an infrared camera, and accounting for the influence of the thin surface layer on the divertor, the heat transfer coefficient of the divertor is determined by solving the heat conduction equation. To benchmark this method, the calculated results are compared with those from ANSYS software under identical conditions (constant temperature boundary and constant heat flux boundary). Additionally, the thermal flux distribution characteristics induced by ELM losses in the EAST divertor are presented.

\subsection{Physics-Informed Neural Network} 
Domain decomposition divides the computational domain into subdomains, each trained with a smaller neural network. This enables parallel computing across subdomains, accelerating computation while efficiently handling high-dimensional and geometrically complex problems~\cite{jagtap2020extended,moseley2023finite,jagtap2020conservative}. Network architecture design significantly enhances the accuracy of PINNs while addressing diverse computational challenges. By utilizing quadratic residual terms, QRes~\cite{bu2021quadratic} enables neural network architectures with higher capacity at each layer to approximate more complex functions with fewer parameters. Based on the Kolmogorov-Arnold representation theorem, which states that any multivariate continuous function can be expressed as a composition and linear combination of univariate functions, KAN~\cite{liu2024kan} provides a promising approach for designing PINNs. PirateNets~\cite{wang2024piratenets} enable stable and efficient utilization of deep neural network structures during training by employing adaptive residual connections and unique initialization schemes. 
For transient problems requiring attention to temporal dependencies, as seen in natural language processing and computer vision, where Mamba~\cite{gu2023mamba} and Transformers~\cite{vaswani2017attention} are extensively employed, network architectures such as PINNsFormer~\cite{zhao2023pinnsformer} and PINNMamba~\cite{xu2025sub} offer an effective solution for handling time-dependent PDEs. Spatial coordinates, as one of the inputs, are often randomly sampled in the domain, failing to capture spatial correlations between sampling points. To address this limitation, methodologies like RoPINN~\cite{wu2024ropinn}, ProPINN~\cite{wu2025propinn}, and Setpinns~\cite{nagda2024setpinns} have been developed to establish spatial relationships within the domain.

Since PINNs require multiple loss terms to decrease simultaneously, and there may be conflicts between these loss terms, if one loss term dominates, it may lead to false convergence, where the loss function decreases but the solution error increases. To address this issue, it is necessary to analyze the convergence from the perspective of loss descent gradients and other relevant factors. Analysis of PINN training problems from a loss perspective targets inherent difficulties~\cite{rathore2024challenges} or stems from the fact that their loss functions usually contain multiple discordant additive terms yielding conflicting gradients~\cite{liu2024config}. Some studies have found that despite very low loss in PINNs, there still exists a significant discrepancy with actual values. To address this issue, inspired by the finite element method, these studies analyzed the problem from the perspective of finite element approaches~\cite{wu2025propinn,nagda2024setpinns,kang2023pixel}.

\subsection{Heat Conduction} 
Deep Reinforcement Learning and Unsupervised Learning have been applied to address thermal conduction-related tasks such as~\cite{smith2021conjugate,beintema2020controlling,hachem2021deep}, however, these efforts have not directly accounted for the fundamental physics of heat transfer problems. PINNs are gaining increasing popularity across diverse engineering fields due to their ability to effectively address real-world problems characterized by noisy data and often partially missing physical information. In PINNs, automatic differentiation (AD) is leveraged to evaluate differential operators without discretization errors, and multi-task learning frameworks are defined to simultaneously fit observed data while respecting the fundamental governing laws of physics. Hennigh et al.~\cite{hennigh2021nvidia} proposed SimNet, an AI-powered multi-physics simulation framework that integrates artificial intelligence with traditional PDE solvers to achieve data-independent real-time simulation capabilities, making it applicable to tasks such as solving temperature fields in heat sinks. Cai et al.~\cite{cai2020heat} pioneered the application of the PINNs framework to heat conduction problems with partially unknown boundary conditions, achieving the inverse identification of boundary conditions by constructing the loss function with integrals of governing PDEs and incorporating sensor measurements. Peng et al.~\cite{peng2024multi} proposed a framework based on PINNs that integrates the heat conduction equation with experimental data to address the challenge of thermal history prediction in multi-layer Directed Energy Deposition (DED) processes, where interlayer thermal accumulation significantly complicates accurate temperature field modeling. Zhang et al.~\cite{zhang2022multi} proposed M-PINN, which handles different material regions with independent PINNs and establishes relationships between material layers using continuity conditions, thereby realizing the solution of steady-state heat conduction problems. Xu et al.~\cite{xu2023physics} trained a physics-informed convolutional encoder-decoder neural network to predict temperature/heat flux fields in various porous media without relying on labeled data.

\section{Our Proposed Approach} 

In this section, we will first give a preliminary introduction to the PINNs and an overview of our proposed HFPINN model. 
Then, we dive into the details of the heat flux estimation problem we study and describe the details about the input encoding, network architecture, and loss functions.

\subsection{Preliminary: Physics-Informed Neural Networks} 

PINNs (Physics-Informed Neural Networks)~\cite{cai2021PINNSurvey, karniadakis2021physics} are a computational paradigm that leverages neural networks to solve partial differential equations (PDEs) or other physics-driven problems. They incorporate physical laws, such as PDEs and boundary conditions, into the loss function of the neural network, enabling the model to satisfy physical constraints while fitting the data.

\begin{figure}
    \centering
    \includegraphics[width=\linewidth]{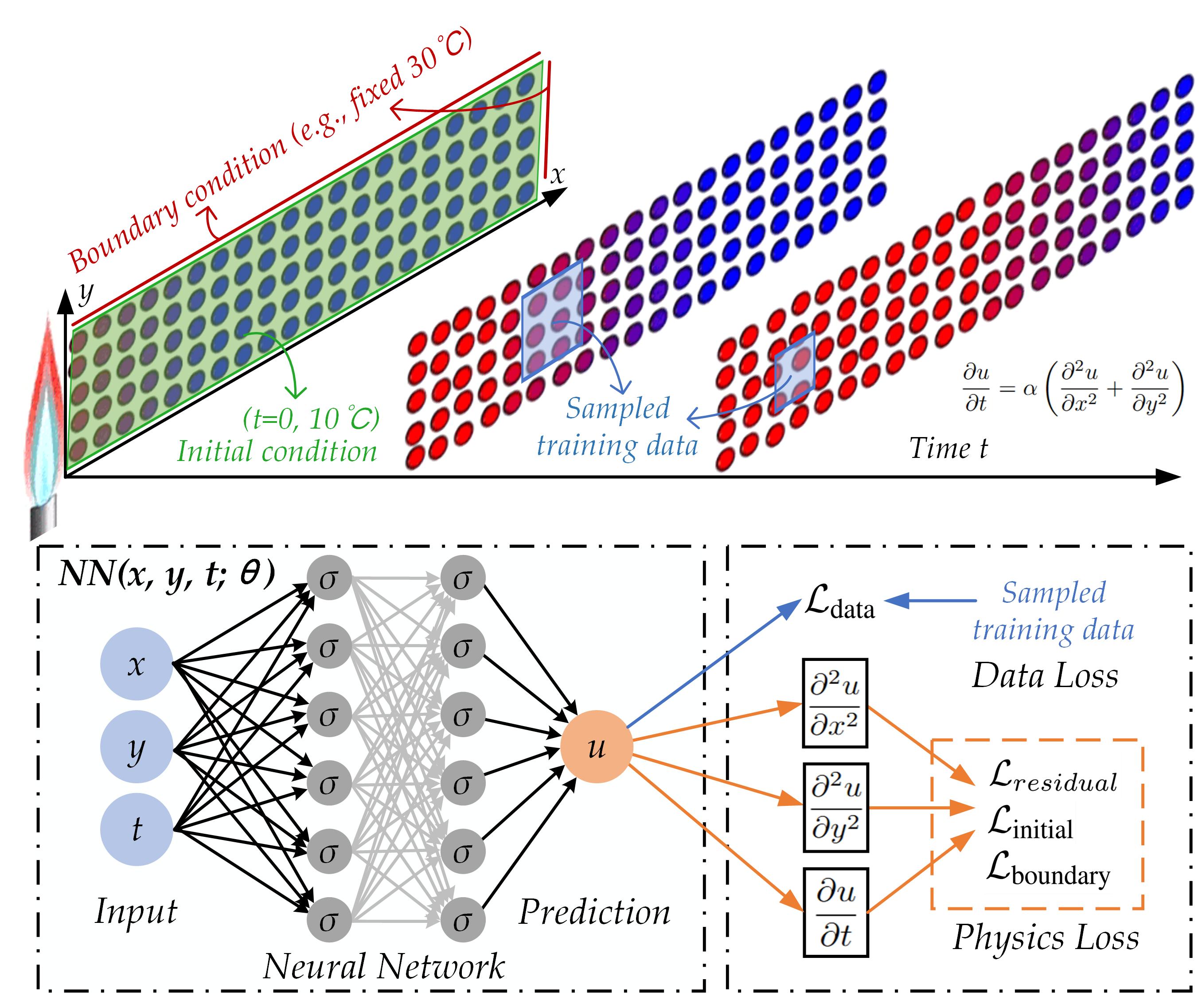}
    \caption{Schematic diagram of solving the 2D transient heat conduction process using PINNs}
    \label{fig:2D_heat_conduction}
\end{figure}

As shown in Fig.~\ref{fig:2D_heat_conduction}, let's take the 2D heat conduction as an example to demonstrate how we can model this process using a PINN. As we know, heat conduction refers to the process by which heat is transferred from high-temperature regions to low-temperature regions within a material. In two-dimensional space, we can describe this process using the two-dimensional heat conduction equation, i.e., 
\begin{equation}
\label{HCPINNs}
\frac{\partial u}{\partial t}=\alpha\left(\frac{\partial^{2} u}{\partial x^{2}}+\frac{\partial^{2} u}{\partial y^{2}}\right)
\end{equation}
where $u(x, y, t)$ represents the temperature distribution, which varies with spatial position $(x, y)$ and time $t$. $\alpha$ is the thermal diffusivity, a physical parameter that describes the heat conduction capability of the material. Actually, this formula is essentially a Partial Differential Equation (PDE) that describes how heat propagates in a two-dimensional plane. If we use the FEM to solve this problem, we need to discretize the continuous space into small elements and approximate the solution of the heat conduction equation on each element. Although this fine-grained solving method is highly accurate, it also has the drawback of high computational cost, making it difficult to achieve fast solutions for real-time problems or large-scale systems.

In contrast, the core idea of using PINNs to model this problem is to leverage neural networks to directly learn the solution of the heat conduction equation while satisfying both physical priors and data constraints. Specifically, we can break it down into the following four steps:  
\textit{Firstly}, assume we have a neural network $f_{\theta}(x, y, t)$, where the inputs are the spatial coordinates $(x, y)$ and time $t$, and the output is the predicted temperature value $u(x, y, t)$. Here, the parameters $\theta$ of the neural networks need to be optimized through training. 
\textit{Secondly}, we know that the heat conduction equation must hold, thus we can calculate whether the neural network output $f_{\theta}(x, y, t)$ satisfies the equation and get the residual loss: 
\begin{equation}
\label{HCPINNs}
\mathcal{L}_{residual} =\frac{\partial f_{\theta}}{\partial t}-\alpha\left(\frac{\partial^{2} f_{\theta}}{\partial x^{2}}+\frac{\partial^{2} f_{\theta}}{\partial y^{2}}\right) 
\end{equation}
\textit{Thirdly}, we enforce boundary conditions, such as fixing the temperature at the edges of the metal plate to a specific value (i.e., the Dirichlet boundary conditions): 
\begin{equation}
u(x, y, t) = g_{D}(x, y, t), \quad \text { for }(x, y) \in \partial \Omega
\end{equation}
where $g_{D}(x, y, t)$ is a given boundary temperature function, $\partial\Omega$ is the boundary of the domain.
The loss for the Dirichlet boundary condition can be defined by calculating the mean squared error (MSE) between the neural network's predicted value $f_\theta(x, y, t)$ and the target value $g_D(x, y, t)$: 
\begin{equation}
\mathcal{L}_{\text{boundary}} = \frac{1}{N_{\text{boundary}}} \sum_{j=1}^{N_{\text{boundary}}} \left( f_\theta(x_j, y_j, t_j) - g_D(x_j, y_j, t_j) \right)^2
\end{equation}
where $N_{\text{boundary}}$ denotes the number of boundary condition sampling points, $(x_j, y_j, t_j)$ denotes the spatial coordinates and time of the boundary condition sampling points. 
If we also consider the initial condition, it describes the state of the system at time $t = 0$. For the heat conduction problem, the initial condition is typically expressed as:
\begin{equation}
u(x, y, 0)=u_{0}(x, y), \quad \text { for }(x, y) \in \Omega, 
\end{equation} 
where $u_{0}(x, y)$ is the given initial temperature distribution. 
The loss for the initial condition can be defined by calculating the MSE between the neural network's predicted value $f_\theta(x, y, 0)$ and the target value $u_0(x, y)$:
\begin{equation}
\mathcal{L}_{\text{initial}} = \frac{1}{N_{\text{init}}} \sum_{i=1}^{N_{\text{init}}} \left( f_\theta(x_i, y_i, 0) - u_0(x_i, y_i) \right)^2
\end{equation}
where $N_{\text{init}}$ represents the number of initial condition sampling points, and $(x_i, y_i)$ represents the spatial coordinates of the initial condition sampling points.

Moreover, if some measurement data is available within the domain, an additional data loss can be incorporated to quantify the mismatch between predictions and observed data.
\begin{equation}
\mathcal{L}_{\text{data}} = \frac{1}{N_{d}} \sum_{i=1}^{N_{d}} 
(f_{\theta}(x_i, y_i, t_i) - {u}_{\text{data}}^{i})^2
\label{eq:data_loss}
\end{equation}
where $N_d$ denotes the number of sampling points $u_{\text{data}}^{i}$ in the measurement data.

The overall loss function can be formulated as: 
\begin{equation}
\mathcal{L}_{\text{total}} = 
\lambda_{\text{1}} \mathcal{L}_{\text{residual}} + \lambda_{\text{2}} \mathcal{L}_{\text{initial}} + \lambda_{\text{3}} \mathcal{L}_{\text{boundary}} + 
\lambda_{\text{4}} \mathcal{L}_{\text{data}} 
\end{equation}
where $\lambda_{\text{1}}, \lambda_{\text{2}}, \lambda_{\text{3}}$ and $\lambda_{\text{4}}$ are tradeoff weights to balance each loss function. 
\textit{Fourthly}, we optimize the parameters of the neural network using gradient descent until the loss function is sufficiently minimized. Therefore, the trained neural network can predict the temperature distribution at any given time, with results that are both consistent with physical laws and experimental data.

\begin{figure*}
\centering
\includegraphics[width=\textwidth]{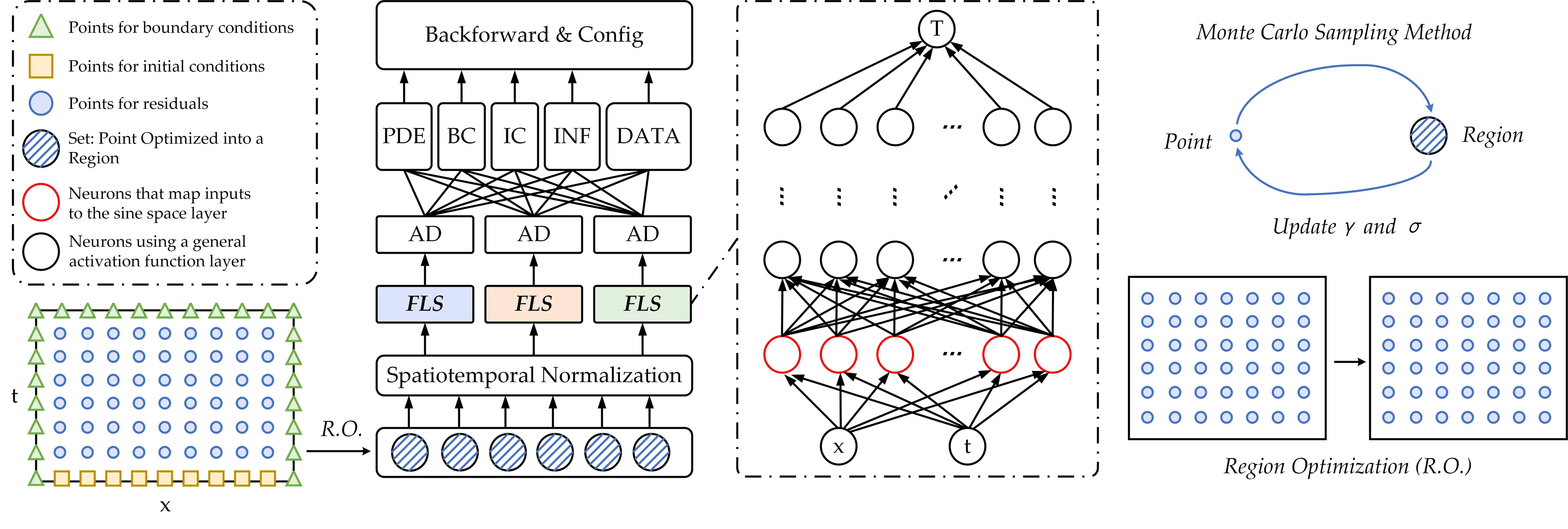} % 使用占位图片
\caption{Our network architecture optimizes sampling points into regions and performs normalization operations. The normalized data is then mapped to the sine domain for processing, and the final result is predicted. During backpropagation (which relies on AD (automatic differentiation)), Config is used to balance various losses.}
\label{fig::framework}
\end{figure*}

\subsection{Overview} 
As shown in Fig.~\ref{fig::framework}, we first generate input data by sampling within the defined spatiotemporal domain of the physical model. Considering that PDEs are defined over continuous space and discrete points may be inadequate for precise solutions across the entire domain, a domain optimization strategy is adopted to extend isolated points into a continuous region. Given significant variations in physical units or numerical ranges of different features, normalization is applied to the input to eliminate dimensional and range discrepancies. We design three fully connected sub-networks to model the corresponding physical materials and resort to the continuity conditions to bridge the inter-network correlations. We also introduce a small amount of supervised data to augment the optimization of our PINNs in addition to the physics-aware loss functions. In summary, by integrating physical equations, we reconstruct the temperature field of the entire EAST divertor water cooling system from sparse data. Extensive experiments in both uniform and non-uniform upper surface scenarios fully validated the effectiveness of our proposed strategy for the heat flux estimation.

\subsection{Problem Formulation }
During EAST discharges, the plasma-facing surface of the divertor is subjected to bombardment by ions at varying temperatures, resulting in non-uniform surface heating. This temperature distribution can be measured using EAST's two sets of wide-field infrared/visible light endoscope systems. The problem we aim to simulate involves reconstructing the overall temperature distribution of the divertor based on the known surface temperature of the divertor.

First of all, we need to model the physical problem to define the spatial domain. Fig.~\ref{fig::firstIMG} (b) shows the physical model of the water-cooled W/Cu monoblock module composed of W, OFHC-Cu, and CuCrZr. The outermost layer is exposed to a steady-state heat flux radiated from the plasma discharge, while the innermost layer is in thermal contact with the coolant. The coolant flows through a cylindrical region with a base radius of 6 mm and a height of 12 mm, maintaining a constant temperature of 22°C. At the interface between the coolant domain and the CuCrZr layer, the convective heat transfer boundary condition is imposed:
\begin{equation}
-k\left(
\frac{\partial T}{\partial x}n_{x} + 
\frac{\partial T}{\partial y}n_{y} + 
\frac{\partial T}{\partial z}n_{z}
\right) = h(T - T_{f})
\label{eq:heat_transfer_boundary}
\end{equation}
where $ k $ is the thermal conductivity of the material, $ n_x, n_y, n_z $ are the unit normal vectors, $ h $ is the convective heat transfer coefficient, and $ T_f $ is the condensate water temperature.

The length, width, and height of the entire device are 30 mm, 28 mm, and 12 mm, respectively. The material in the hollow cylindrical region with inner and outer radius of 6 mm and 7.5 mm at the bottom and a height of 12 mm is CuCrZr. The material in the hollow cylindrical region with inner and outer radii of 7.5 mm and 10.5 mm at the bottom and a height of 12 mm is Cu. The material in the remaining regions is W. The heat conduction equation is expressed in the Cartesian coordinate system as:
\begin{equation}
\frac{\partial}{\partial x}\left(k\frac{\partial T}{\partial x}\right) + 
\frac{\partial}{\partial y}\left(k\frac{\partial T}{\partial y}\right) + 
\frac{\partial}{\partial z}\left(k\frac{\partial T}{\partial z}\right) = 
\rho c_{p}\frac{\partial T}{\partial t}
\label{eq:heat_conduction_basic}
\end{equation}
where $k$ is the thermal conductivity of the material, $\rho$ is the density of the material, and $c_p$ is the specific heat capacity of the material. Material properties are listed in Table~\ref{tab:thermal_properties}.

\begin{table}
\centering
\caption{Information on thermal properties of the target material.}
\label{tab:thermal_properties}
\begin{tabular}{l|ccc} 
\hline
Material & $ k \, (\text{W/m·K}) $ & $ \rho \, (\text{kg/m}^3) $ & $ c_p \, (\text{J/kg·K}) $ \\
\hline
W        & 173                      & 19298                   & 129                     \\
Cu       & 403                      & 8960                    & 390                     \\
CuCrZr   & 318                      & 8920                    & 388                     \\
\hline
\end{tabular}
\end{table}

To simulate the surface temperature changes induced by particle impact and facilitate explanation, the temperature distribution on the upper surface is set as a constant-temperature boundary condition, satisfying:
\begin{equation}
T = 300 \times e^{-\frac{(x - 0.014)^2}{0.006 \times 0.006}}
\label{eq:gaussian_temp}
\end{equation}
Except for the upper surface, all other surfaces satisfy the adiabatic boundary condition, expressed as:
\begin{equation}
\frac{\partial T}{\partial n} = 0
\label{eq:neumann_basic}
\end{equation}
where $n$ denotes the boundary normal vector.

For transient thermal simulations, initial conditions are essential to provide a well-defined starting point for spatiotemporal evolution, ensuring temporal causality and physical plausibility of the solution. To provide an initial condition for the transient heat conduction problem, we set the initial temperature condition to 30°C.

\subsection{Input Representation}
Considering that the original unit of the input coordinates was meters, their excessively small order of magnitude ($10^{-3}$) made it difficult for the neural network to effectively distinguish the differences between input coordinates, ultimately causing the network's prediction results to fall into a trivial solution (i.e., all predicted results converged to the same temperature value). To enable the network to discern input differences, we performed normalization processing on the input coordinates:
\begin{equation}
\begin{cases}
\hat{x} = 2\left( \frac{x - x_{\min}}{x_{\max} - x_{\min}} \right) - 1 \\
\hat{y} = 2\left( \frac{y - y_{\min}}{y_{\max} - y_{\min}} \right) - 1 \\
\hat{z} = 2\left( \frac{z - z_{\min}}{z_{\max} - z_{\min}} \right) - 1 \\
\hat{t} = 2\left( \frac{t - t_{\min}}{t_{\max} - t_{\min}} \right) - 1
\end{cases}
\label{eq:neumann_basic}
\end{equation}
Here, $\hat{x}, \hat{y}, \hat{z}, \hat{t}$ represent the normalized results, while $x_{\min},y_{\min},z_{\min},t_{\min}$ represent the minimum values in the training data, and $x_{\max},y_{\max},z_{\max},t_{\max}$ represent the maximum values in the training data.
By normalizing the input data to the range $[-1, 1]$, not only is the convergence rate accelerated, but also the loss decrease becomes more stable.

Due to the limitations of numerical computation, PINNs are traditionally optimized at a limited number of selected points. However, since PDEs are generally defined over continuous domains, optimizing the model solely at scattered points may be insufficient to obtain accurate solutions over the entire domain. To address this inherent limitation of default scattered optimization, this paper adopts a regional optimization training paradigm. A similar region optimization strategy is used in other PINNs~\cite{wu2024ropinn}.  
\begin{equation}
\mathcal{L}_r^\mathrm{region}(u_\theta, \mathcal{S}) = 
\frac{1}{|\mathcal{S}|} \sum_{x \in \mathcal{S}} \mathcal{L}_r^\mathrm{region}(u_\theta, x) 
\end{equation}
where $u_\theta$ represents the neural network parameterized by $\theta$, and $\mathcal{S}$ denotes the finite set of selected points.

For the $t$-th iteration process, a Monte Carlo approximation is used to sample from the neighborhood of sampling points, expressed as:
\begin{equation}
\mathcal{S}' = \left\{ \boldsymbol{x}_i + \boldsymbol{\xi}_i \right\}_{i=1}^{|\mathcal{S}|}, \quad \boldsymbol{x}_i \in \mathcal{S}, \quad \boldsymbol{\xi}_i \sim U\left[ 0, \frac{r}{\sigma_t} \right]^{d+1}
\label{eq:neumann_basic}
\end{equation}
where $r$ denotes the default region size and $\sigma_t$ is the trust region calibration value. $U$ is the uniform multivariate distribution, and $\mathcal{S}$ is the original set of points. By recording the gradient parameter $g_t$ in each optimization round, the initially empty gradient recording buffer $\mathbf{g}$ is updated, and subsequently $\sigma$ is updated, which is expressed as:  
\begin{equation}
\sigma_t =||\sigma(\mathbf{g})||
\label{eq:neumann_basic}
\end{equation}
where $|| * ||$ denotes the $L1$ norm.

\subsection{Network Architecture}
Drawing on the domain-decomposed PINNs~\cite{jagtap2020extended, moseley2023finite, jagtap2020conservative} and the network architecture used in~\cite{zhang2022multi} for solving multi-medium 2D steady-state heat conduction, we extend it to address higher-dimensional transient heat conduction problems. Each distinct material region is addressed by an independent PINN, which is designed to predict the temperature within the corresponding region of the material. Given that the interface between adjacent materials must satisfy consistency in temperature and heat flux, the interface conditions are formulated as:
\begin{equation}
T_{Ni} = T_{Nj}
\label{eq:neumann_basic}
\end{equation}
\begin{equation}
k_i\frac{\partial T_{Ni}}{\partial n} = k_j\frac{\partial T_{Nj}}{\partial n}
\label{eq:neumann_basic}
\end{equation}
where $k$ denotes thermal conductivity, $n$ denotes the boundary normal vector and $i = j-1$ represents adjacent regions of different materials. Through the continuity conditions at the interface, the independent networks in different domains can be connected, and simultaneously the boundary conditions outside the local domain can be perceived.

% 这一段有点乱，要修改一下[这里要修改一下]
As shown in Fig.~\ref{fig::framework}, our network consists of three sub-networks interconnected via interface conditions. For each sub-network, we first project it into a sine space, followed by approximating the temperature field $T$ using a fully connected neural network. By training the network parameters $\theta$ to minimize the loss function and leveraging AD to compute spatiotemporal derivatives, we ensure the accurate representation of the PDEs. 
The network architecture comprises three key components, i.e., a normalization layer, which eliminates the impact of dimensional units while accelerating model convergence; a region optimization layer, which expands isolated points to their continuous domain to reduce generalization error; a fully connected network with sine mapping is responsible for fitting temperature values. Different material domains are coupled at their interfaces, with imposed interface conditions ensuring the conservation of temperature and heat flux.

The loss functions of many learning problems contain multiple additional terms, which may be inconsistent and generate conflicting update directions. We adopt the method proposed in~\cite{liu2024config}, which ensures positive dot products between the final update and the loss-specific gradients of each term to provide conflict-free updates. This approach also maintains a consistent optimization rate for all loss terms and dynamically adjusts the gradient magnitudes based on the conflict level.

\begin{figure*}
    \centering
    \includegraphics[width=0.9\linewidth]{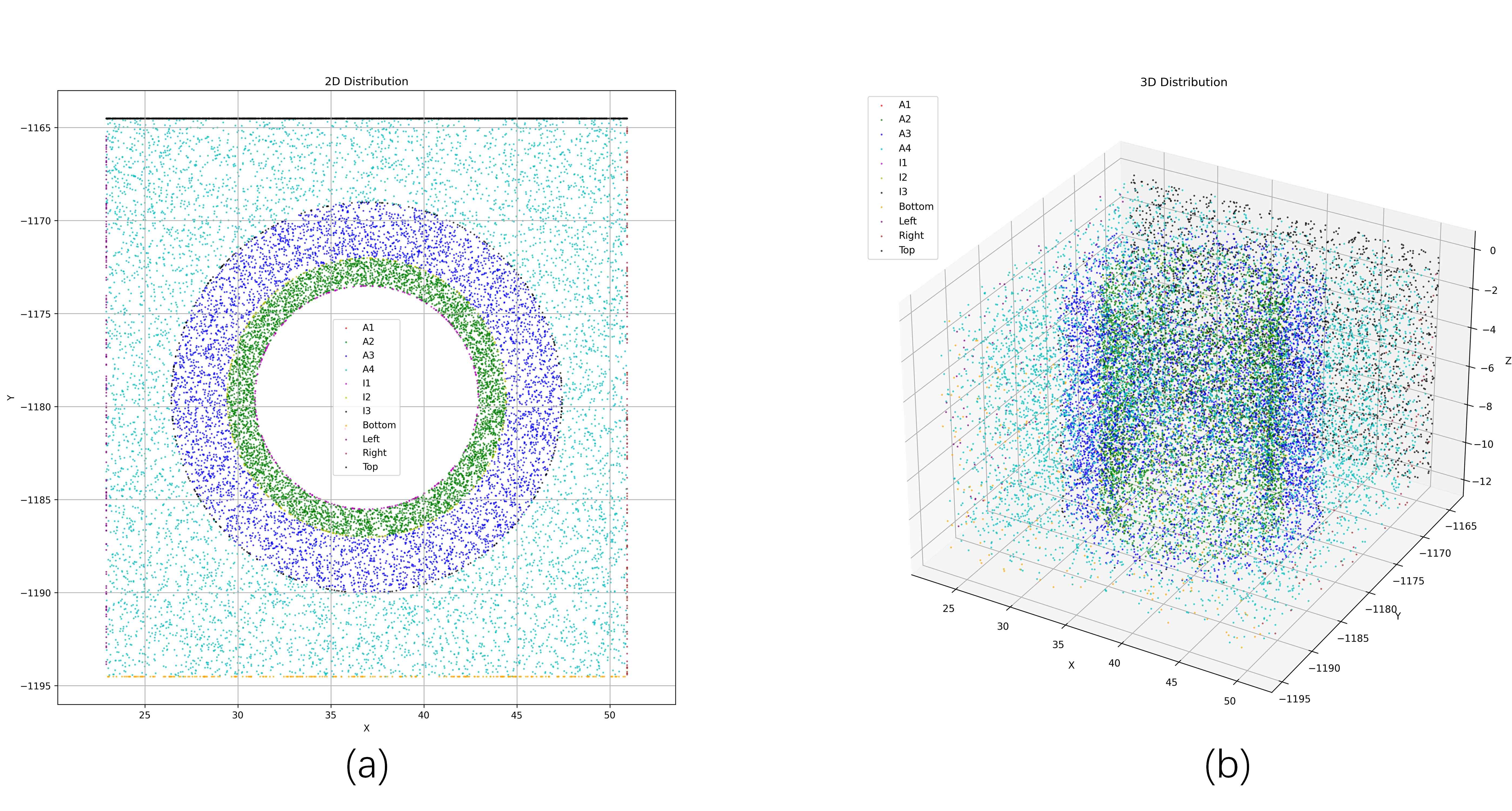}
    \caption{(a) Two-dimensional (2D) sampling distribution. (b) Three-dimensional (3D) sampling distribution.}
    \label{fig:sample}
\end{figure*}
\subsection{Loss Function}

The loss functions used for the optimization of the PINNs are described below. 
Specifically, constant temperature loss is used to enable the temperature field to fit and perceive the heat source, expressed as:
\begin{equation}
\mathcal{L}_{Constant} = ||T - T_{constant}||^2
\label{eq:neumann_basic}
\end{equation}
Here, $||\ *\ ||^2$ represents the $L2$ distance, and $T_{constant}$ denotes the constant temperature at the upper surface.

Adiabatic loss ensures that there is no heat inflow or outflow at the boundary under this condition, expressed as:
\begin{equation}
\mathcal{L}_{\text{Adiabatic}} = ||\frac{\partial{T}}{\partial{n}}||^2
\label{eq:neumann_basic}
\end{equation}
Here, $n$ represents the normal vector of the corresponding boundary.

Convective heat loss enables the heat arriving at the surface to be carried away by the condensate water, expressed as:
\begin{equation}
\mathcal{L}_{\text{Convective}} =|| -k \frac{\partial T}{\partial n} - h(T - T_f)||^2
\label{eq:neumann_basic}
\end{equation}
Here, $k$ represents the thermal conductivity, $h$ denotes the convective heat transfer coefficient, and $T_{f}$ indicates the coolant (water) temperature.

Conductive heat loss is used to constrain the temperature field to satisfy the heat conduction equation, expressed as:
\begin{equation}
\mathcal{L}_{\text{Heat}} = ||\rho c_p \frac{\partial u}{\partial t} 
                - \frac{\partial}{\partial x}\left(k \frac{\partial u}{\partial x}\right) 
                - \frac{\partial}{\partial y}\left(k \frac{\partial u}{\partial y}\right) 
                - \frac{\partial}{\partial z}\left(k \frac{\partial u}{\partial z}\right)||^2
\label{eq:neumann_basic}
\end{equation}
Here, $\rho, c_p, and \  k$ respectively represent the density, specific heat capacity, and thermal conductivity of the corresponding material.

Temperature consistency loss is used to constrain that the predicted temperatures at the interface of adjacent networks should be consistent, expressed as:
\begin{equation}
\mathcal{L}_{\text{Consistency}}=||T_{Ni}-T_{Nj}||^2
\label{eq:neumann_basic}
\end{equation}

Heat flux consistency loss ensures that energy transfer conforms to physical laws, expressed as:
\begin{equation}
\mathcal{L}_{\text{Flux}}=||k_i\frac{\partial T_{Ni}}{\partial n} - k_j\frac{\partial T_{Nj}}{\partial n}||^2
\label{eq:neumann_basic}
\end{equation}
Here, $N_i$ and $N_j$ represent the adjacent networks, while $k_i$ and $k_j$ denote the thermal conductivities of the materials corresponding to their respective network regions.

Initial loss provides a starting point for transient problems, expressed as:
\begin{equation}
\mathcal{L}_{\text{Init}}=||T-T_{init}||^2
\label{eq:neumann_basic}
\end{equation}
Here, $T_{init}$ represents the initial temperature.

To mitigate the network from falling into trivial solutions, we introduce a small amount of data as supervision, and the data loss is expressed as:
\begin{equation}
\mathcal{L}_{\text{DATA}}=||T-T_{DATA}||^2
\label{eq:neumann_basic}
\end{equation}

To balance the various losses, we sum all the losses and assign weights to them to form the total loss, expressed as:
\begin{equation}
\small 
\begin{split}
\mathcal{L}_{\text{total}} = \lambda_{1}\mathcal{L}_{Constant} + \lambda_{2}\mathcal{L}_{Adiabatic} + \lambda_{3}\mathcal{L}_{Convective} + \lambda_{4}\mathcal{L}_{Heat} \\
+ \lambda_{5}\mathcal{L}_{Consistency} + \lambda_{6}\mathcal{L}_{Flux} + \lambda_{7}\mathcal{L}_{Init} + \lambda_{8}\mathcal{L}_{DATA}
\end{split}
\label{eq:loss_total}
\end{equation}
where $\lambda_1$ to $\lambda_8$ represent the tradeoff weights used to balance each loss function.

\begin{figure*}[!t] % [!t] 表示尝试放置于页面顶部
    \centering
    \includegraphics[width=\textwidth]{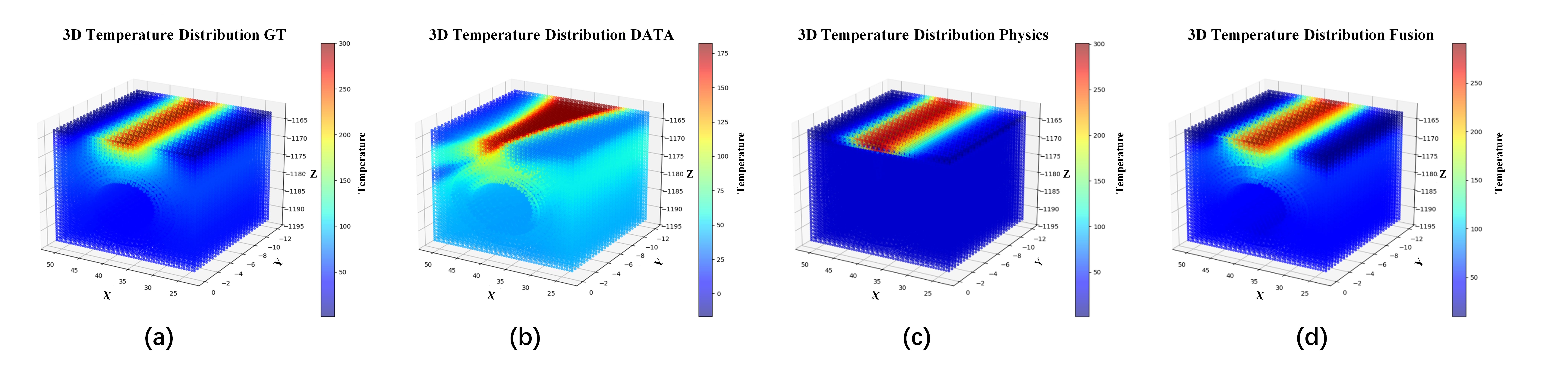} % 使用占位图片
    \caption{Comparative analysis of physical constraints, data-driven methods, and their combination: (a) Finite element simulation, (b) Data-driven, (c) Physical constraints, (d) Combination of data and physical equations.}
    \label{fig.3}
\end{figure*}

\section{Experiments}

\subsection{Dataset and Evaluation Metric} 

\noindent \textbf{Dataset}. We investigated two sets of experiments: one where the upper surface temperature was maintained at a constant 100°C, as shown in Fig.~\ref{fig:Alaysis} (b), and another where the upper surface temperature followed a Gaussian distribution to simulate the uneven temperature distribution caused by particle bombardment on the upper surface, as shown in Fig.~\ref{fig:Alaysis} (a).  
As shown in Fig.~\ref{fig:sample}, for experiments where the upper surface is maintained at 100°C, the data sources are divided into sampling data and supervised data. 
For sampling data, we adopted a random sampling strategy, specifically, for different material regions, 1845 points in the CuCrZr region, 5160 points in the Cu region, and 6500 points in the W region were sampled. 
For boundaries adhering to adiabatic boundary conditions, 1000 points were sampled; for boundaries under constant temperature boundary conditions, 2500 points were sampled; and for regions with convection heat transfer boundaries, 250 points were collected. 
Additionally, 170 points were sampled at the interface between CuCrZr and Cu, and 290 points were sampled at the interface between Cu and W. 
For the supervised data, we randomly sampled 5, 10, and 15 data points from the CuCrZr, Cu, and W regions, respectively, out of a total of 496,330 points simulated using the finite element method. 
For experiments where the upper surface temperature follows a Gaussian distribution, the sampling points were consistent with those in the previous experiment. 
For the supervised data, we increased the number of sampled points in the W, Cu, and CuCrZr regions to 30, 50, and 80, respectively. 

\noindent \textbf{Evaluation Metric}. To evaluate the performance of the models, we take relative Mean Absolute Error (rMAE), relative Root Mean Square Error (rRMSE) following common practice~\cite{zhao2023pinnsformer}, and Mean Absolute Error (MAE). The metrics are formulated as: 
\begin{equation} 
\text{rMAE}(\hat{u}) = \frac{\sum_{n=1}^N |\hat{u}(x_n, t_n) - u(x_n, t_n)|}{\sum_{n=1}^N |u(x_n, t_n)|},
\end{equation}
\begin{equation} 
\text{rRMSE}(\hat{u}) = \sqrt{\frac{\sum_{n=1}^N |\hat{u}(x_n, t_n) - u(x_n, t_n)|^2}{\sum_{n=1}^N |u(x_n, t_n)|^2}}, 
\end{equation}
\begin{equation} 
\text{MAE}(\hat{u}) = \frac{1}{N} \sum_{n=1}^N |\hat{u}(x_n, t_n) - u(x_n, t_n)|
\end{equation}
where $ N $ is the number of test points, $ u(x, t) $ is the ground-truth solution, and $ \hat{u}(x, t) $ is the model's prediction.

\subsection{Implementation Details} 
For our method, we train our network using the Adam optimizer with an initial learning rate of 0.001 and the CosineAnnealingLR learning rate scheduler. Each network architecture comprises 4 layers, with 256 neurons configured per layer. As shown in Eq.~\ref{eq:loss_total}, the weights of the loss terms $[\lambda_{1}, \lambda_{2}, \lambda_{3},\lambda_{4}, \lambda_{5}, \lambda_{6}, \lambda_{7}, \lambda_{8}]$ are set to [1, 1, 1, 1, 5, 1, 1, 10]. 
All experiments were implemented in PyTorch 2.0.0 and trained on an NVIDIA 3090 GPU. More details can be found in our source code.

\begin{table}
\small % 缩小字体
\setlength{\tabcolsep}{3pt} % 减小列间距
\centering
\caption{Component analysis of the key modules in our HFPINNs. 
Specifically, DATA: impact of data guidance, MD: region decomposition, Set: Region Optimization, Config: gradient processing, and Select: sampling point selection}
\label{tab:Ablation}
\begin{tabular}{ccccccccc} % 调整列类型，前5列为c（中心对齐），后3列为S（数值对齐）
\toprule
PINN & MD & DATA & Set & Config & Select & {rMAE} & {rRMSE} & {MAE} \\
\midrule
 & $\checkmark$ & $\checkmark$ & &$\checkmark$ & & 0.119 & 0.259 & 7.38 \\
$\checkmark$ & $\checkmark$ & $\checkmark$ & & & & 0.063 & 0.118 & 3.93 \\
$\checkmark$ & & $\checkmark$ & & $\checkmark$&  &  0.557 & 0.709 & 34.60 \\
$\checkmark$ & $\checkmark$ & $\checkmark$ & & $\checkmark$ & & 0.058 &  0.106 & 3.62 \\
$\checkmark$ & $\checkmark$ & $\checkmark$ & $\checkmark$ & $\checkmark$ & & 0.056 & 0.131 & 3.52\\
$\checkmark$ & $\checkmark$ & $\checkmark$ & $\checkmark$ & $\checkmark$ & $\checkmark$ & 0.053 & 0.086& 3.31 \\
\bottomrule
\end{tabular}
\end{table}

\begin{figure*}
    \centering
    \includegraphics[width=\textwidth]{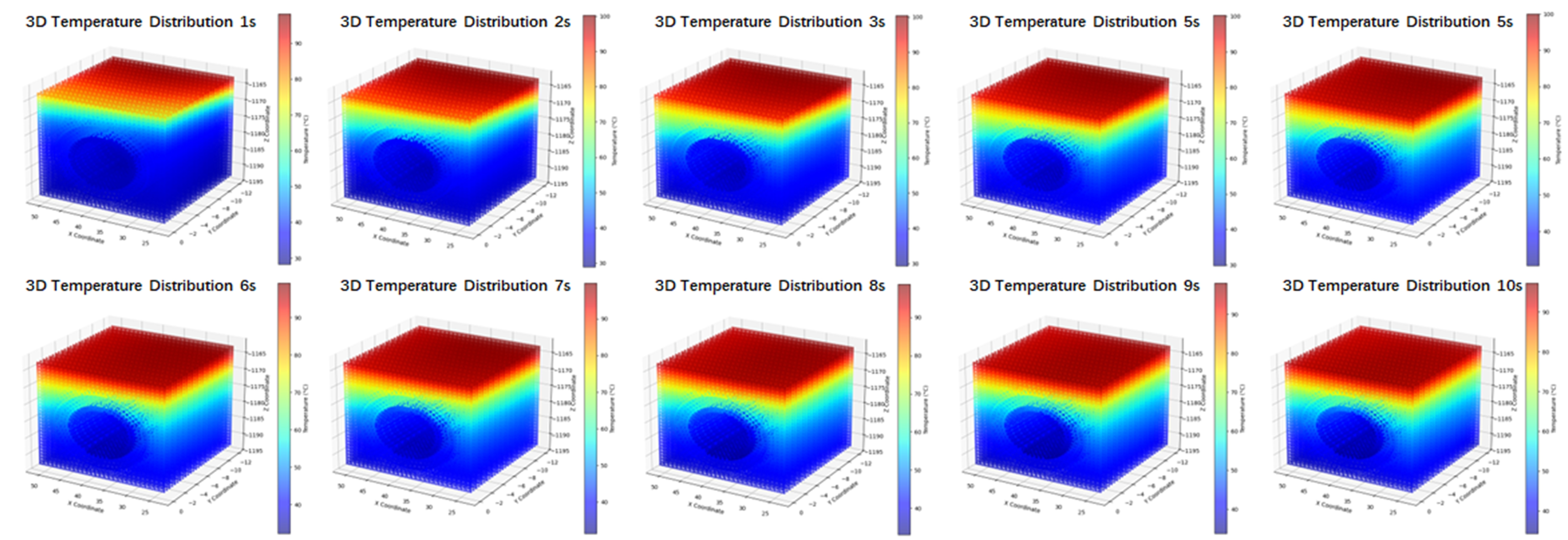} % 使用占位图片
    \caption{Temperature variation with time under constant temperature conditions.} 
    \label{fig.4}
\end{figure*}

\subsection{Component Analysis }
As shown in Table~\ref{tab:Ablation}, we evaluated the impact of data guidance, region decomposition, gradient processing, and sampling point selection on the final prediction results of the network. 

\noindent \textbf{Data guidance (DATA)}. If we train the network using only a small amount of data, although the mean absolute error is relatively low, the training results fail to satisfy physical constraints, resulting in an uneven temperature distribution. Conversely, if we rely solely on physical equations as constraints, while the temperature distribution becomes uniform, most regions fall into trivial solutions. Introducing a small amount of data as guidance can significantly enhance the accuracy of both approaches while ensuring the temperature distribution conforms to physical laws, as shown in Fig.~\ref{fig.3}.

\noindent \textbf{Region decomposition (MD)}. If only a single network is used to predict the entire temperature field, the loss may fail to decrease properly due to the incomplete consistency of physical laws across different regions. By decomposing the domain into sub-networks based on materials, the computational cost for each sub-region is significantly reduced. Meanwhile, the sub-regions exchange and fuse information through interface conditions, enabling faster and more accurate acquisition of solutions for the entire domain.

\noindent \textbf{Gradient processing (Config)}. Loss terms associated with initial/boundary conditions and physical equations have received particular attention. By ensuring that the dot product between the final update and the gradient specific to each loss is positive, conflict-free updates are provided, which reduces conflicts between losses and thereby improves performance.

\noindent \textbf{Sampling point selection (Select)}. Experimental findings reveal that in the temperature field, the region near the heat source (i.e., the upper surface of the model) exhibits significant temperature variations and is also the area with the largest errors. Increasing the sampling ratio in this region can effectively improve the accuracy of temperature prediction.

\subsection{Ablation Study}

\begin{table}
\centering         % 表格水平居中显示
\caption{Comparison of Performance Metrics of Different Activation Functions Under Varying Temperature Conditions}  % 表格标题（显示在表格上方）
\label{tab:activation_performance}     % 表格标签（用于文中引用，如 \ref{tab:activation_performance}）
\begin{tabular}{l c c c}  % 列对齐方式：左对齐（方法列），右对齐（数值列）
    \toprule          % 顶部粗线（来自 booktabs 宏包）
    \textbf{Method} & \textbf{rMAE} & \textbf{rRMSE} & \textbf{MAE} \\  % 表头
    \midrule          % 中间细线
    ReLU & 0.216 & 0.348 & 13.419\\
    Swish & 0.053 & 0.095 & 3.309\\
    Tanh & 0.147 & 0.068 & 4.25 \\
    GELU & 0.060 & 0.097 & 3.724\\
    Wave & 0.077 & 0.100 & 4.810 \\
    \bottomrule       % 底部粗线
\end{tabular}
\end{table}

\begin{figure*}[!htp] % 尝试将图片置于页面顶部（跨双栏）
    \centering % 图片整体居中
    \includegraphics[width=\textwidth]{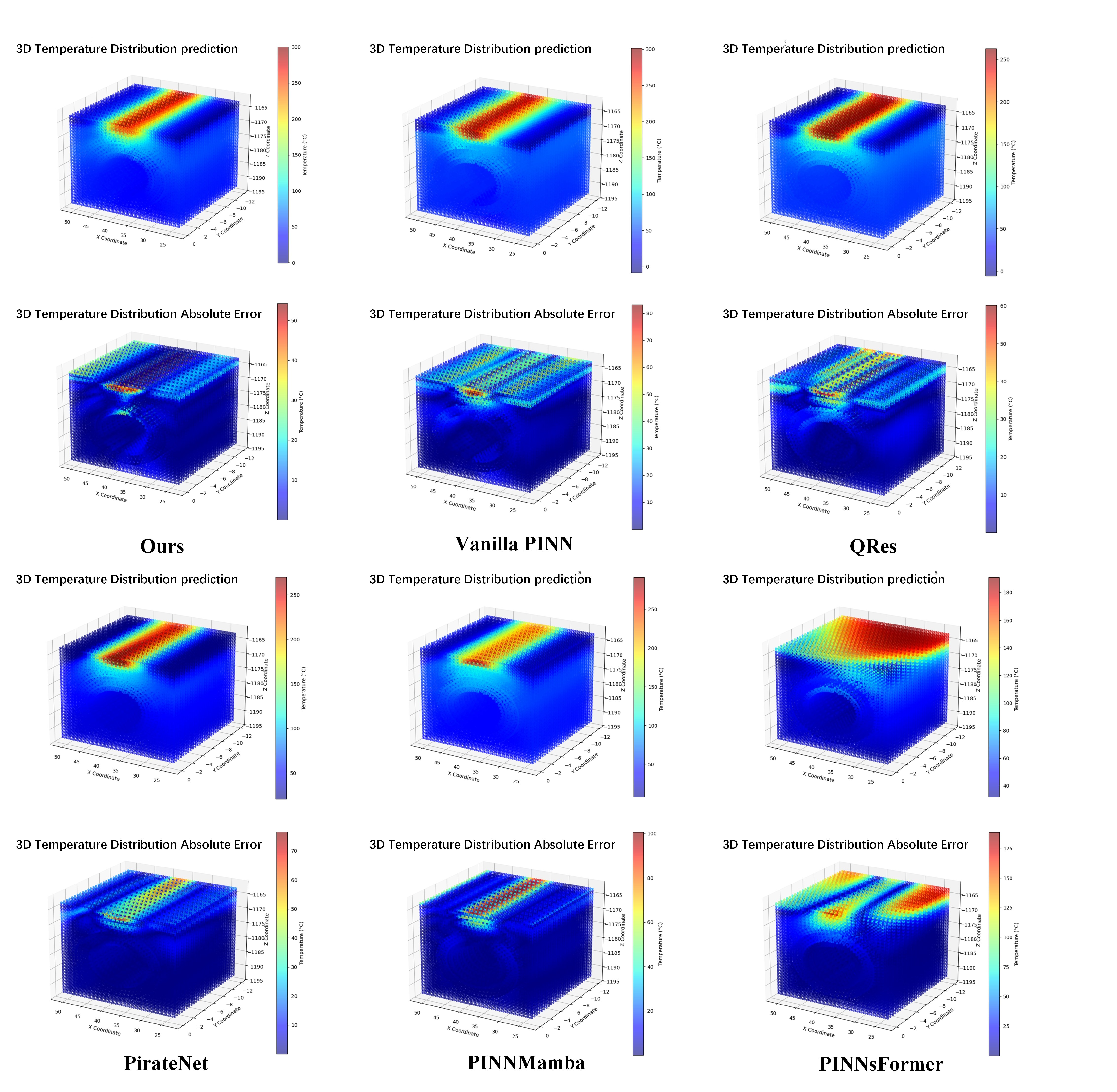} % 缩放图片至原宽度的 80%
    \caption{Under different network frameworks, a comparison of prediction results and losses is presented. For each network framework, the upper part shows the prediction results, and the lower part shows the mean absolute error.} % 文字说明居中
    \label{fig.5}
\end{figure*}

\noindent $\bullet$ \textbf{Analysis of different activation functions.~}
Table~\ref{tab:activation_performance} demonstrates the impact of different activation functions on the experimental results under the Vanilla PINN framework. ReLU is non-differentiable at the origin, leading to discontinuous gradient propagation and making it difficult to capture continuous physical field variations. While Wave activation functions perform well for periodically varying functions, the heat conduction equation lacks periodic characteristics. Given the large temperature span in heat conduction problems, Tanh may cause increased absolute errors. Tab.~\ref{tab:activation_performance100} presents the experimental results under constant temperature conditions.

\begin{table}
\centering   
\caption{Comparison of Performance Metrics of Different Activation Functions Under Constant Temperature Conditions}  % 表格标题（显示在表格上方）
\label{tab:activation_performance100} 
\begin{tabular}{l r r r}  % 列对齐方式：左对齐（方法列），右对齐（数值列）
    \toprule          % 顶部粗线（来自 booktabs 宏包）
    \textbf{Method} & \textbf{rMAE} & \textbf{rRMSE} & \textbf{MAE} \\  % 表头
    \midrule          % 中间细线
    ReLU & 0.044 & 0.052 & 2.52\\
    Swish & 0.042 & 0.057 & 2.40\\
    Tanh & 0.047 & 0.060 & 2.70  \\
    GELU & 0.048 & 0.071 & 2.79\\
    Wave & 0.058 & 0.081 & 3.31 \\
    \bottomrule       % 底部粗线
\end{tabular}
\end{table}

\begin{table}[ht]  % [ht] 表示表格位置（此处/顶部）
\centering         % 表格居中显示
\caption{Comparison of Different Network Architectures Under Varying Temperature Conditions}  % 表格标题
\label{tab:network_performance}       % 表格标签（用于文中引用）
\begin{tabular}{l r r r}  % 列对齐方式：左对齐（方法列），右对齐（数值列）
    \toprule
    \textbf{Network Architecture} & \textbf{rMAE} & \textbf{rRMSE} & \textbf{MAE} \\
    \midrule
    Vanilla PINN~\cite{raissi2019physics} & 0.147 & 0.068 & 4.25 \\
    QRes~\cite{bu2021quadratic}   & 0.060 & 0.114 & 3.773 \\
    PirateNet~\cite{wang2024piratenets} & 0.079 & 0.128 & 4.905 \\
    PINNsFormer~\cite{zhao2023pinnsformer} & 0.222 & 0.355 & 13.832 \\
    PINNMamba~\cite{xu2025sub}    & 0.078 & 0.154 & 4.849 \\
    \midrule
    Ours   & 0.056 & 0.131 & 3.522 \\
    \bottomrule
\end{tabular}
\end{table}

\noindent $\bullet$ \textbf{Analysis on different network architecture.~} 
Our method uses FLS to map inputs to a sine space for processing, which reduces spectral deviation. Although QRes achieves comparable accuracy to our method, its training time is significantly longer. Despite the excellent performance of PINNsFormer and PINNMamba on the Convection, Reaction, and Wave equations, they exhibit poor performance on three-dimensional heat conduction problems. Fig.~\ref{fig.5} shows the experimental results under different network architectures and compares them with FEM results. The impact of different network architectures on experimental results is shown in Table~\ref{tab:network_performance}, and Table~\ref{tab:network_performance100} shows the experimental results under constant temperature conditions.

% 暂时存放
\begin{table}[!htp]
\centering   
\caption{Comparison of Different Network Architectures Under Constant Temperature Conditions}  % 表格标题
\label{tab:network_performance100}       % 表格标签（用于文中引用）
\begin{tabular}{l r r r}  % 列对齐方式：左对齐（方法列），右对齐（数值列）
    \toprule
    \textbf{Network Architecture} & \textbf{rMAE} & \textbf{rRMSE} & \textbf{MAE} \\
    \midrule
    Vanilla PINN~\cite{raissi2019physics} & 0.047 & 0.060 & 2.70 \\
    QRes~\cite{bu2021quadratic}   & 0.045 & 0.058 & 2.59 \\
    PirateNet~\cite{wang2024piratenets} & 0.049 & 0.063 & 2.84\\
    PINNsFormer~\cite{zhao2023pinnsformer} & 0.077 & 0.098 & 4.39 \\
    PINNMAmba~\cite{xu2025sub}    & 0.037 & 0.051 & 2.13 \\
    \midrule
    Ours   & 0.043 & 0.059 & 2.50 \\
    \bottomrule
\end{tabular}
\end{table}

\begin{figure*}
    \centering
    \includegraphics[width=1\textwidth]{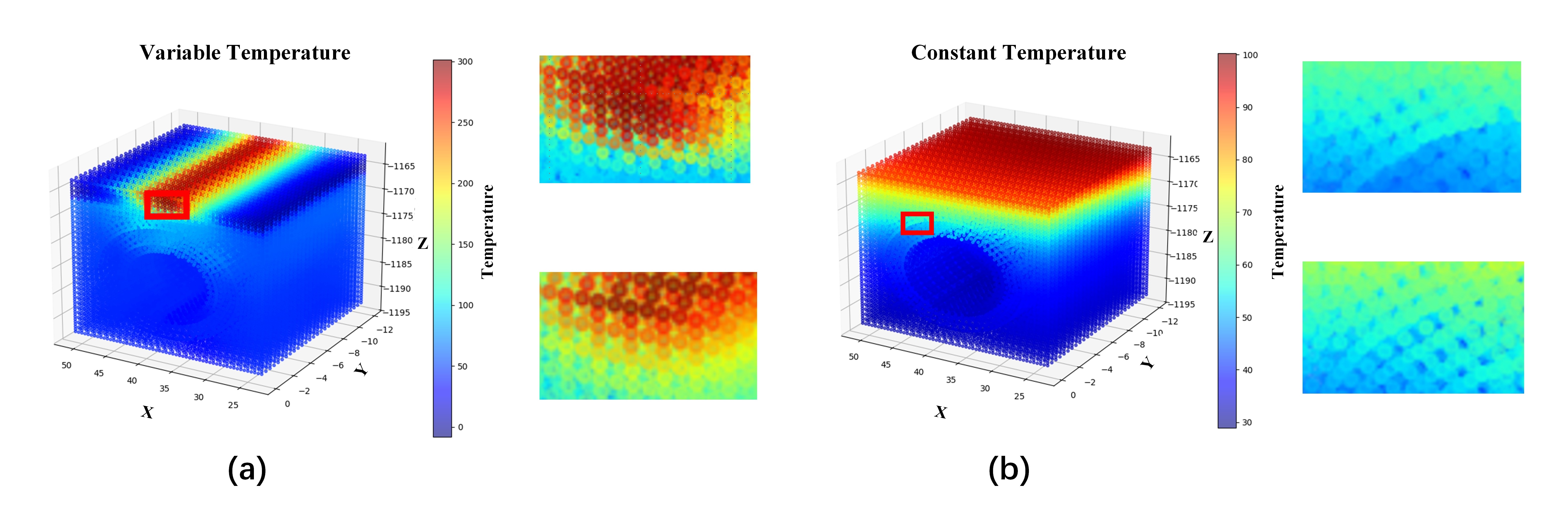}
    \caption{Regions with large fitting errors under constant temperature conditions. The top-right corner shows the predicted results, and the bottom-right corner shows the finite element simulation results.}
    \label{fig:Alaysis}
\end{figure*}

\begin{figure}[!htp]
    \centering
    \includegraphics[width=1\linewidth]{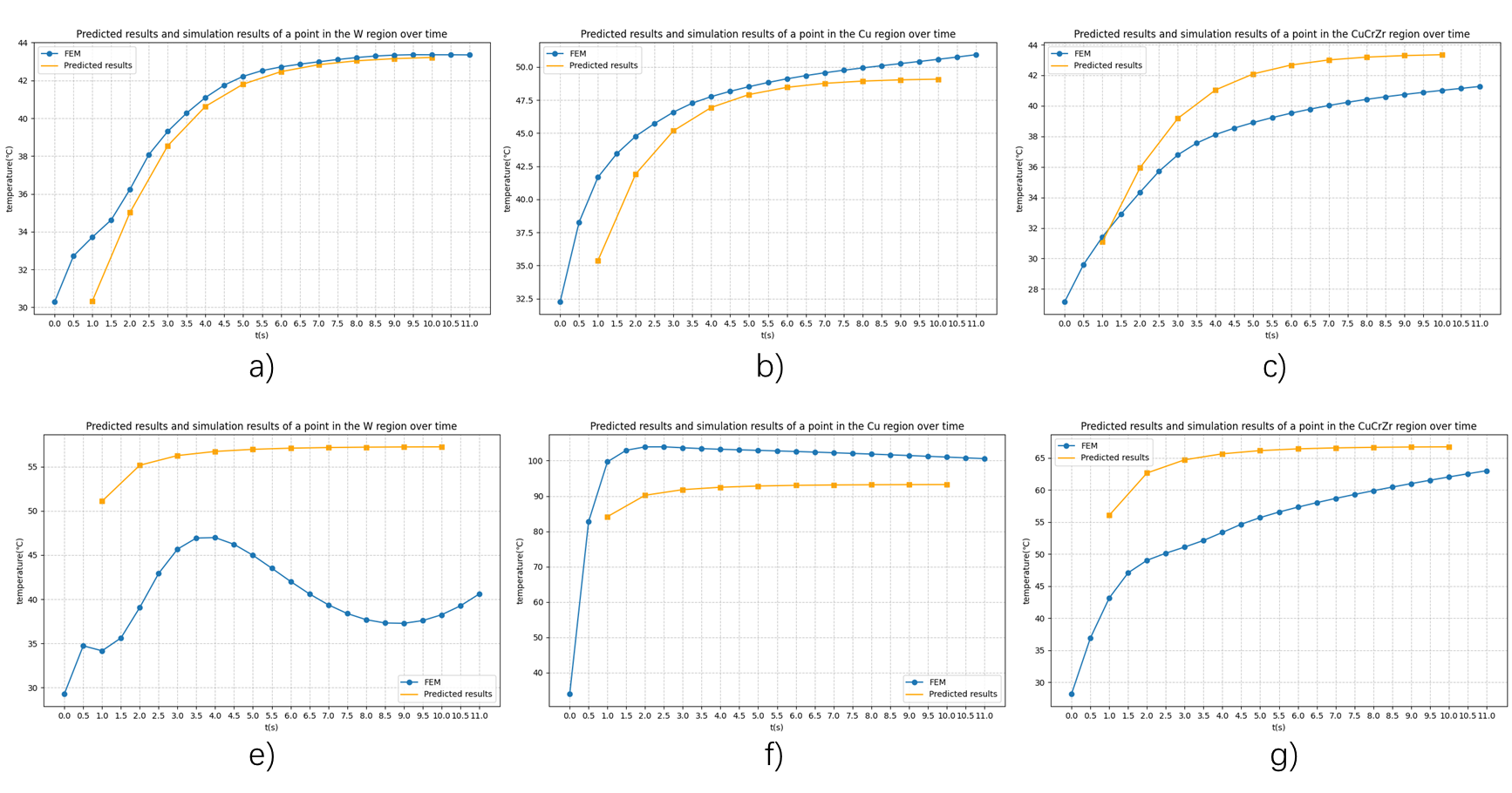}
    \caption{Predicted and simulated temperatures of different material regions over time: Positive samples a) W region; b) Cu region; c) CuCrZr region;
     Negative samples d) W region; e) Cu region; f) CuCrZr region.}
    \label{fig.6}
\end{figure}

\begin{figure}[!htp]
    \centering
    \includegraphics[width=1\linewidth]{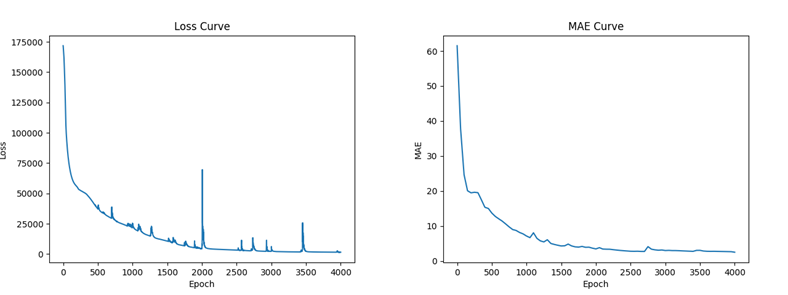}
    \caption{Plot of the decreasing curves of loss and MAE with the iteration process}
    \label{fig:7}
\end{figure}

% \begin{figure}[!htp]
%     \centering
%     \includegraphics[width=1\linewidth]{figures/image9.png}
%     \caption{Regions with large fitting errors under variable temperature conditions. The top-right corner shows the predicted results, and the bottom-right corner shows the finite element simulation results.}
%     \label{fig:9}
% \end{figure}

\noindent $\bullet$ \textbf{Analysis on Errors.~} 
Fig.~\ref{fig.6} shows the temperature variation with time at different sampling points under variable temperature conditions. Within the figure, points with better performance and points with poorer performance were selected from different regions. It was found that points with better performance are mainly distributed in regions with lower temperatures, and the CuCrZr point is farther from the water-cooled boundary condition. Points with poorer performance tend to be distributed in high-temperature regions. In other words, high-temperature regions are often key areas requiring focused sampling. Meanwhile, Fig.~\ref{fig.4} shows the predicted temperature variation over time under constant temperature conditions.  Fig.~\ref{fig:7} illustrates the variation of the L2 loss and Mean Absolute Error as the number of training epochs increases. As training progresses, we observe that MAE does not consistently decrease as loss decreases; instead, it exhibits a trend of first decreasing and then increasing. In the initial stage of training, the network prioritizes fitting the data loss. When the data loss drops to a certain threshold, the network begins to emphasize equation constraints more strongly. Consequently, MAE decreases under the combined constraints of data and equations. However, as loss continues to decrease further, under the condition of a set constant temperature of 100°C, MAE rises and tends toward a trivial solution. Under variable temperature conditions, the convective heat transfer loss fails to decrease further, causing MAE to fluctuate around the minimum value. If a smaller weight is assigned to the continuity loss, although convective heat transfer loss may decrease, MAE will increase as the loss continues to decrease beyond the previous minimum value. This phenomenon may arise because the simulated data generated by the finite element method (used for comparison) differs from real-world data, leading to an imbalance between data loss and control equation loss. 
Fig.~\ref{fig:Alaysis} shows the regions with large fitting errors under constant and variable temperature conditions.

\subsection{Efficiency Analysis} 
Since the FEM cannot independently compute a single second and must first calculate all time steps, a specific second can then be extracted. For the finite element method, we take the average prediction time per frame (each frame contains 5,573 grid points) as the inference time. We also sample 5,573 points as inference inputs for the PINNs. As shown in Table~\ref{tab:Efficiency}. It can be observed that once the network is trained, the PINN is nearly 40 times faster than the FEM in prediction.

\begin{table}[!htp]
\centering   
\caption{Efficiency Comparison Between FEM and PINNs}  % 表格标题（显示在表格上方）
\label{tab:Efficiency} 
\begin{tabular}{c c c} % l=左对齐(方法列), r=右对齐(数值列)
    \toprule
    \textbf{Method} & \textbf{Training Time} & \textbf{Reasoning Time} \\
    \midrule
    FEM & / & 0.02s \\
    HFPINN & 43min19s & 0.00054s \\
    \bottomrule
\end{tabular}
\end{table}

\subsection{Limitation Analysis} 
We employed the PINNs method to develop a surrogate model for solving the heat conduction problem in the EAST divertor. However, our model requires a small amount of finite element data for training, as without data, the prediction results tend to converge to a trivial solution. This limitation restricts our ability to reconstruct solutions in data-scarce scenarios. Second, although our model achieves a small average temperature difference, the maximum temperature discrepancy with results from the FEM remains notable at specific points. Finally, our model requires retraining for different temperature conditions, indicating relatively poor generalization capability.

\section{Conclusion} 
The proposed multi-domain decomposition PINN architecture, through the integration of physical constraints and data-driven methods, achieves a significant improvement in efficiency (with an inference speed nearly 40 times faster than FEM) for the three-dimensional transient heat conduction problem in the EAST divertor. Additionally, training strategy optimizations have led to a certain enhancement in accuracy compared to traditional PINNs. Although errors in high-temperature regions and at material interfaces still require further resolution, its high efficiency demonstrates its potential application value in real-time monitoring of fusion engineering.

% \section*{Acknowledgment} 

% 放置绘制的图和表，最后确定放置位置

\small{ 
\bibliographystyle{IEEEtran}
\bibliography{reference}

\begin{thebibliography}{10}
\providecommand{\url}[1]{#1}
\csname url@samestyle\endcsname
\providecommand{\newblock}{\relax}
\providecommand{\bibinfo}[2]{#2}
\providecommand{\BIBentrySTDinterwordspacing}{\spaceskip=0pt\relax}
\providecommand{\BIBentryALTinterwordstretchfactor}{4}
\providecommand{\BIBentryALTinterwordspacing}{\spaceskip=\fontdimen2\font plus
\BIBentryALTinterwordstretchfactor\fontdimen3\font minus
  \fontdimen4\font\relax}
\providecommand{\BIBforeignlanguage}[2]{{%
\expandafter\ifx\csname l@#1\endcsname\relax
\typeout{** WARNING: IEEEtran.bst: No hyphenation pattern has been}%
\typeout{** loaded for the language `#1'. Using the pattern for}%
\typeout{** the default language instead.}%
\else
\language=\csname l@#1\endcsname
\fi
#2}}
\providecommand{\BIBdecl}{\relax}
\BIBdecl

\bibitem{wang2024multi}
S.~Wang, Y.~Wang, Q.~Ma, X.~Wang, N.~Yan, Q.~Yang, G.~Xu, and J.~Tang,
  ``Multi-modal fusion based q-distribution prediction for controlled nuclear
  fusion,'' in \emph{International Conference on Brain Inspired Cognitive
  Systems}.\hskip 1em plus 0.5em minus 0.4em\relax Springer, 2024, pp.
  115--125.

\bibitem{ma2024exploiting}
Q.~Ma, S.~Wang, T.~Zheng, X.~Dai, Y.~Wang, Q.~Yang, and X.~Wang, ``Exploiting
  memory-aware q-distribution prediction for nuclear fusion via modern hopfield
  network,'' in \emph{International Conference on Brain Inspired Cognitive
  Systems}.\hskip 1em plus 0.5em minus 0.4em\relax Springer, 2024, pp.
  104--114.

\bibitem{yang2020FEMHEATEAST}
Z.~Yang, P.~He, H.~Yan, Z.~Bin, S.~Shuangbao, F.~Wang, M.~Chen, G.~Jia, and
  X.~Gong, ``The development of a three-dimensional finite element method code
  for the heat flux analysis of tungsten monoblock divertor on east,''
  \emph{Fusion Engineering and Design}, vol. 152, p. 111448, 2020.

\bibitem{shi2017heat}
B.~Shi, Z.-D. Yang, B.~Zhang, C.~Yang, K.-F. Gan, M.-W. Chen, J.-H. Yang,
  H.~Zhang, J.-L. Qi, X.-Z. Gong \emph{et~al.}, ``Heat flux on east divertor
  plate in h-mode with lhcd/lhcd+ nbi,'' \emph{Chinese Physics Letters},
  vol.~34, no.~9, p. 095201, 2017.

\bibitem{shi2018study}
B.~Shi, C.~Yang, Z.~Yang, D.~Cheng, H.~Wang, J.~Yang, H.~Zhang, J.~Qi,
  Q.~Zhang, X.~Gong \emph{et~al.}, ``Study of temperature and heat flux on the
  east divertor target plate in lhw+ nbi/icrh h-mode,'' \emph{IEEE Transactions
  on Plasma Science}, vol.~46, no.~7, pp. 2672--2676, 2018.

\bibitem{chen2023development}
G.~Chen, P.~Cao, J.~Yang, R.~Liang, L.~Li, Y.~Sun, and F.~Zhong, ``Development
  of a high-speed small-angle infrared thermography system in east,''
  \emph{Review of Scientific Instruments}, vol.~94, no.~5, 2023.

\bibitem{wang2023MMPTMSurvey}
X.~Wang, G.~Chen, G.~Qian, P.~Gao, X.-Y. Wei, Y.~Wang, Y.~Tian, and W.~Gao,
  ``Large-scale multi-modal pre-trained models: A comprehensive survey,''
  \emph{Machine Intelligence Research}, vol.~20, no.~4, pp. 447--482, 2023.

\bibitem{wang2025xihefusion}
X.~Wang, Q.~Yang, F.~Wang, Q.~Chen, W.~Wu, Y.~Jin, J.~Jiang, L.~Jin, B.~Jiang,
  D.~Sun \emph{et~al.}, ``Xihefusion: Harnessing large language models for
  science communication in nuclear fusion,'' \emph{arXiv preprint
  arXiv:2502.05615}, 2025.

\bibitem{raissi2019physics}
M.~Raissi, P.~Perdikaris, and G.~E. Karniadakis, ``Physics-informed neural
  networks: A deep learning framework for solving forward and inverse problems
  involving nonlinear partial differential equations,'' \emph{Journal of
  Computational physics}, vol. 378, pp. 686--707, 2019.

\bibitem{wang2024physics}
F.~Wang, Z.~Zhai, Z.~Zhao, Y.~Di, and X.~Chen, ``Physics-informed neural
  network for lithium-ion battery degradation stable modeling and prognosis,''
  \emph{Nature Communications}, vol.~15, no.~1, p. 4332, 2024.

\bibitem{jarolim2023probing}
R.~Jarolim, J.~Thalmann, A.~Veronig, and T.~Podladchikova, ``Probing the solar
  coronal magnetic field with physics-informed neural networks,'' \emph{Nature
  Astronomy}, vol.~7, no.~10, pp. 1171--1179, 2023.

\bibitem{jang2024grad}
B.~Jang, A.~A. Kaptanoglu, R.~Gaur, S.~Pan, M.~Landreman, and W.~Dorland,
  ``Grad--shafranov equilibria via data-free physics informed neural
  networks,'' \emph{Physics of Plasmas}, vol.~31, no.~3, 2024.

\bibitem{cai2021PINNSurvey}
S.~Cai, Z.~Mao, Z.~Wang, M.~Yin, and G.~E. Karniadakis, ``Physics-informed
  neural networks (pinns) for fluid mechanics: A review,'' \emph{Acta Mechanica
  Sinica}, vol.~37, no.~12, pp. 1727--1738, 2021.

\bibitem{karniadakis2021physics}
G.~E. Karniadakis, I.~G. Kevrekidis, L.~Lu, P.~Perdikaris, S.~Wang, and
  L.~Yang, ``Physics-informed machine learning,'' \emph{Nature Reviews
  Physics}, vol.~3, no.~6, pp. 422--440, 2021.

\bibitem{causon2010introductory}
D.~Causon and C.~Mingham, \emph{Introductory finite difference methods for
  PDEs}.\hskip 1em plus 0.5em minus 0.4em\relax Bookboon, 2010.

\bibitem{reddy1993introduction}
J.~N. Reddy, ``An introduction to the finite element method,'' \emph{New York},
  vol.~27, no.~14, 1993.

\bibitem{eymard2000finite}
R.~Eymard, T.~Gallou{\"e}t, and R.~Herbin, ``Finite volume methods,''
  \emph{Handbook of numerical analysis}, vol.~7, pp. 713--1018, 2000.

\bibitem{brunner2011comparison}
D.~Brunner, B.~LaBombard, J.~Payne, and J.~Terry, ``Comparison of heat flux
  measurements by ir thermography and probes in the alcator c-mod divertor,''
  \emph{Journal of nuclear materials}, vol. 415, no.~1, pp. S375--S378, 2011.

\bibitem{wan2017overview}
Y.~Wan, J.~Li, Y.~Liu, X.~Wang, V.~Chan, C.~Chen, X.~Duan, P.~Fu, X.~Gao,
  K.~Feng \emph{et~al.}, ``Overview of the present progress and activities on
  the cfetr,'' \emph{Nuclear Fusion}, vol.~57, no.~10, p. 102009, 2017.

\bibitem{jagtap2020extended}
A.~D. Jagtap and G.~E. Karniadakis, ``Extended physics-informed neural networks
  (xpinns): A generalized space-time domain decomposition based deep learning
  framework for nonlinear partial differential equations,''
  \emph{Communications in Computational Physics}, vol.~28, no.~5, 2020.

\bibitem{moseley2023finite}
B.~Moseley, A.~Markham, and T.~Nissen-Meyer, ``Finite basis physics-informed
  neural networks (fbpinns): a scalable domain decomposition approach for
  solving differential equations,'' \emph{Advances in Computational
  Mathematics}, vol.~49, no.~4, p.~62, 2023.

\bibitem{jagtap2020conservative}
A.~D. Jagtap, E.~Kharazmi, and G.~E. Karniadakis, ``Conservative
  physics-informed neural networks on discrete domains for conservation laws:
  Applications to forward and inverse problems,'' \emph{Computer Methods in
  Applied Mechanics and Engineering}, vol. 365, p. 113028, 2020.

\bibitem{bu2021quadratic}
J.~Bu and A.~Karpatne, ``Quadratic residual networks: A new class of neural
  networks for solving forward and inverse problems in physics involving
  pdes,'' in \emph{Proceedings of the 2021 SIAM International Conference on
  Data Mining (SDM)}.\hskip 1em plus 0.5em minus 0.4em\relax SIAM, 2021, pp.
  675--683.

\bibitem{liu2024kan}
Z.~Liu, Y.~Wang, S.~Vaidya, F.~Ruehle, J.~Halverson, M.~Solja{\v{c}}i{\'c},
  T.~Y. Hou, and M.~Tegmark, ``Kan: Kolmogorov-arnold networks,'' \emph{arXiv
  preprint arXiv:2404.19756}, 2024.

\bibitem{wang2024piratenets}
S.~Wang, B.~Li, Y.~Chen, and P.~Perdikaris, ``Piratenets: Physics-informed deep
  learning with residual adaptive networks,'' \emph{Journal of Machine Learning
  Research}, vol.~25, no. 402, pp. 1--51, 2024.

\bibitem{gu2023mamba}
A.~Gu and T.~Dao, ``Mamba: Linear-time sequence modeling with selective state
  spaces,'' \emph{arXiv preprint arXiv:2312.00752}, 2023.

\bibitem{vaswani2017attention}
A.~Vaswani, N.~Shazeer, N.~Parmar, J.~Uszkoreit, L.~Jones, A.~N. Gomez,
  {\L}.~Kaiser, and I.~Polosukhin, ``Attention is all you need,''
  \emph{Advances in neural information processing systems}, vol.~30, 2017.

\bibitem{zhao2023pinnsformer}
Z.~Zhao, X.~Ding, and B.~A. Prakash, ``Pinnsformer: A transformer-based
  framework for physics-informed neural networks,'' \emph{arXiv preprint
  arXiv:2307.11833}, 2023.

\bibitem{xu2025sub}
C.~Xu, D.~Liu, Y.~Hu, J.~Li, R.~Qin, Q.~Zheng, and J.~Xiong, ``Sub-sequential
  physics-informed learning with state space model,'' \emph{arXiv preprint
  arXiv:2502.00318}, 2025.

\bibitem{wu2024ropinn}
H.~Wu, H.~Luo, Y.~Ma, J.~Wang, and M.~Long, ``Ropinn: Region optimized
  physics-informed neural networks,'' \emph{arXiv preprint arXiv:2405.14369},
  2024.

\bibitem{wu2025propinn}
H.~Wu, Y.~Ma, H.~Zhou, H.~Weng, J.~Wang, and M.~Long, ``Propinn: Demystifying
  propagation failures in physics-informed neural networks,'' \emph{arXiv
  preprint arXiv:2502.00803}, 2025.

\bibitem{nagda2024setpinns}
M.~Nagda, P.~Ostheimer, T.~Specht, F.~Rhein, F.~Jirasek, S.~Mandt, M.~Kloft,
  and S.~Fellenz, ``Setpinns: Set-based physics-informed neural networks,''
  \emph{arXiv preprint arXiv:2409.20206}, 2024.

\bibitem{rathore2024challenges}
P.~Rathore, W.~Lei, Z.~Frangella, L.~Lu, and M.~Udell, ``Challenges in training
  pinns: A loss landscape perspective,'' \emph{arXiv preprint
  arXiv:2402.01868}, 2024.

\bibitem{liu2024config}
Q.~Liu, M.~Chu, and N.~Thuerey, ``Config: Towards conflict-free training of
  physics informed neural networks,'' \emph{arXiv preprint arXiv:2408.11104},
  2024.

\bibitem{kang2023pixel}
N.~Kang, B.~Lee, Y.~Hong, S.-B. Yun, and E.~Park, ``Pixel: Physics-informed
  cell representations for fast and accurate pde solvers,'' in
  \emph{Proceedings of the AAAI conference on artificial intelligence},
  vol.~37, no.~7, 2023, pp. 8186--8194.

\bibitem{smith2021conjugate}
R.~Smith and S.~Dutta, ``Conjugate thermal optimization with unsupervised
  machine learning,'' \emph{Journal of Heat Transfer}, vol. 143, no.~5, p.
  052901, 2021.

\bibitem{beintema2020controlling}
G.~Beintema, A.~Corbetta, L.~Biferale, and F.~Toschi, ``Controlling
  rayleigh--b{\'e}nard convection via reinforcement learning,'' \emph{Journal
  of Turbulence}, vol.~21, no. 9-10, pp. 585--605, 2020.

\bibitem{hachem2021deep}
E.~Hachem, H.~Ghraieb, J.~Viquerat, A.~Larcher, and P.~Meliga, ``Deep
  reinforcement learning for the control of conjugate heat transfer,''
  \emph{Journal of Computational Physics}, vol. 436, p. 110317, 2021.

\bibitem{hennigh2021nvidia}
O.~Hennigh, S.~Narasimhan, M.~A. Nabian, A.~Subramaniam, K.~Tangsali, Z.~Fang,
  M.~Rietmann, W.~Byeon, and S.~Choudhry, ``Nvidia simnet™: An ai-accelerated
  multi-physics simulation framework,'' in \emph{International conference on
  computational science}.\hskip 1em plus 0.5em minus 0.4em\relax Springer,
  2021, pp. 447--461.

\bibitem{cai2020heat}
S.~Cai, Z.~Wang, C.~Chryssostomidis, and G.~E. Karniadakis, ``Heat transfer
  prediction with unknown thermal boundary conditions using physics-informed
  neural networks,'' in \emph{Fluids engineering division summer meeting}, vol.
  83730.\hskip 1em plus 0.5em minus 0.4em\relax American Society of Mechanical
  Engineers, 2020, p. V003T05A054.

\bibitem{peng2024multi}
B.~Peng and A.~Panesar, ``Multi-layer thermal simulation using physics-informed
  neural network,'' \emph{Additive Manufacturing}, vol.~95, p. 104498, 2024.

\bibitem{zhang2022multi}
B.~Zhang, G.~Wu, Y.~Gu, X.~Wang, and F.~Wang, ``Multi-domain physics-informed
  neural network for solving forward and inverse problems of steady-state heat
  conduction in multilayer media,'' \emph{Physics of Fluids}, vol.~34, no.~11,
  2022.

\bibitem{xu2023physics}
J.~Xu, H.~Wei, and H.~Bao, ``Physics-informed neural networks for studying heat
  transfer in porous media,'' \emph{International Journal of Heat and Mass
  Transfer}, vol. 217, p. 124671, 2023.

\end{thebibliography}
}
\end{document}